\documentclass[journal]{IEEEtran}

\ifCLASSINFOpdf
\else
\fi

\usepackage{amssymb}            
\usepackage{mathtools}          
\usepackage{mathrsfs}           
\mathtoolsset{showonlyrefs}     
\usepackage{graphicx}           
\usepackage{subcaption}         
\usepackage[space]{grffile}     
\usepackage{url}                
\usepackage{algorithm}          
\usepackage{algpseudocode}      
\usepackage{float}              
\usepackage{multirow}           
\usepackage{ulem} 
\usepackage{xcolor}

\begin{document}

\title{PEPA: a Persistently Autonomous Embodied Agent with Personalities}

\author{
Kaige Liu,
Yang Li,
Lijun Zhu,
Weinan Zhang%
\thanks{$^{1}$Kaige Liu is with the School of Artificial Intelligence and Automation, Huazhong University of Science and Technology, Wuhan 430074, China, and also with Shanghai Innovation Institute, Shanghai 200231, China
        {\tt\footnotesize kgliu@hust.edu.cn}}%
\thanks{$^{2}$Yang Li is with the School of Computer Science,
Shanghai Jiao Tong University, Shanghai 200240, China. {\tt\footnotesize yang.li.cs@sjtu.edu.cn}}%
\thanks{$^{3}$Lijun Zhu is with the School of Artificial Intelligence and Automation,
Huazhong University of Science and Technology, Wuhan 430074, China, and
also with the State Key Laboratory of Intelligent Manufacturing Equipment
and Technology, Huazhong University of Science and Technology, Wuhan
430074, China
        {\tt\footnotesize ljzhu@hust.edu.cn}}%
\thanks{$^{4}$Weinan Zhang is with the School of Computer Science,
Shanghai Jiao Tong University, Shanghai 200240, China, and also with Shanghai Innovation Institute, Shanghai 200231, China
        {\tt\footnotesize wnzhang@sjtu.edu.cn}}%
\thanks{Weinan Zhang and Yang Li are the corresponding authors.}    
}


\maketitle

\begin{abstract}
Living organisms exhibit persistent autonomy through internally generated goals and self-sustaining behavioral organization, yet current embodied agents remain driven by externally scripted objectives. 
This dependence on predefined task specifications limits their capacity for long-term deployment in dynamic, unstructured environments where continuous human intervention is impractical.
We propose that personality traits provide an intrinsic organizational principle for achieving persistent autonomy. 
Analogous to genotypic biases shaping biological behavioral tendencies, personalities enable agents to autonomously generate goals and sustain behavioral evolution without external supervision.
To realize this, we develop PEPA, a three-layer cognitive architecture that operates through three interacting systems: 
Sys3 autonomously synthesizes personality-aligned goals and refines them via episodic memory and daily self-reflection; 
Sys2 performs deliberative reasoning to translate goals into executable action plans; 
Sys1 grounds the agent in sensorimotor interaction, executing actions and recording experiences. 
This closed loop of goal generation, execution, memory consolidation, and reflection enables the agent to continuously redefine objectives and adapt behavior over extended operational periods.
We validate the framework through real-world deployment on a quadruped robot in a multi-floor office building. 
Operating without reliance on fixed task specifications, the robot autonomously arbitrates between user requests and personality-driven motivations, navigating elevators and exploring environments accordingly.
Quantitative analysis across five distinct personality prototypes demonstrates stable, trait-aligned behaviors. 
The results confirm that personality-driven cognitive architectures enable sustained autonomous operation characteristic of persistent embodied systems.
Code and demos are available at \url{https://sites.google.com/view/pepa-persistent/}.
\end{abstract}


\IEEEpeerreviewmaketitle

\section{Introduction}
Living organisms achieve persistent autonomy \cite{Zeng2015autonomy} through intrinsic behavioral organization, sustaining self-directed operation over extended periods without external instruction. 
This capacity is increasingly critical for robotic systems deployed in real-world environments where continuous human oversight is impractical. 
An eldercare companion \cite{Chen2019Companionship} is expected to autonomously decide when to initiate social interaction, when to monitor health indicators, and when to allow privacy based on evolving patterns in the resident's behavior. 
A planetary exploration \cite{Hassanalian2018EvolutionOS} rover encountering unexpected terrain features have to autonomously determine whether to investigate anomalies or continue planned traversal, balancing scientific opportunity against resource constraints without waiting for instructions from Earth. 
These scenarios pose challenges that go beyond high-performance on predefined benchmarks, because real-world operation is open-ended, non-stationary, and constrained by physical resources such as energy, wear, and risk \cite{kunze2018longterm,hawes2016strands}. 
These applications demand what we term persistent autonomy: the capacity for agents to autonomously generate goals, pursue them coherently over extended periods, and adapt behavior based on intrinsic organizational principles rather than external supervision.

An embodied agent is grounded in a sensorimotor loop, where competence arises from continuous coupling between body and environment \cite{brooks1991intelligence,beer1995dynamical}.
Embodiment constrains and enables autonomy through real dynamics and resources, shaping cognition \cite{clark1997being} and offloading part of control to physical morphology \cite{pfeifer2007self}. On this basis, achieving persistent autonomy in embodied agents requires three capabilities beyond episodic task execution. 
First, agents must maintain continuous functionality under physical constraints: managing energy, avoiding damage, and ensuring safety in uncontrolled environments \cite{kunze2018longterm,Liu2021ALL}. 
Second, they must accumulate and reorganize knowledge under non-stationary conditions, adapting to evolving environmental dynamics and internal state changes \cite{meng2025preserving,kirkpatrick2017overcoming}.
Third, and most critically, they must autonomously generate behavioral objectives when external goals are absent, deciding what to do, when to do it, and how to prioritize among competing opportunities and risks. 
This third capability represents the fundamental challenge: without external task specification, what determines an agent's goals? How does it maintain behavioral coherence across extended operational periods while remaining adaptive to new experiences? This is not merely a control or learning problem, but an organizational one requiring intrinsic principles to structure long-horizon behavior.

Current approaches fail to address this organizational challenge. 
Recent progress in learning-based control and large language models has enabled impressive task competence \cite{ouyang2025long,wang2024survey}, including persistent agent frameworks such as Sophia \cite{sun2025sophia}, Reflexion \cite{shinn2024reflexion}, and Evolving Agents \cite{li2024evolving}, which introduce lifetime goal hierarchies, self-reflection, or evolving personality traits. Embodied evolution explores population-level adaptation \cite{Bredche2017EmbodiedEI,kanagawa2024evolution}. However, these systems remain either externally goal-driven or largely disembodied and unconstrained by physical survival and resource limits \cite{wang2024survey}. Long-term autonomy and lifelong learning focus on robustness and knowledge retention \cite{kunze2018longterm,kirkpatrick2017overcoming,meng2025preserving}, yet assume prescribed objectives. 
Most deployed robots still operate under externally scripted goals and fixed reward templates \cite{arkin1998behavior,bredeche2018embodied}. Personality modeling in social robotics, often grounded in the Big Five \cite{mcadams1992five} and extended with affective or memory-based mechanisms \cite{picard1997affective,churamani2020affect,tang2025robot}, demonstrates socially distinguishable behaviors but treats personality as a static design parameter rather than an intrinsic principle for autonomous goal generation. 
Collectively, these approaches lack the intrinsic behavioral organization required for truly persistent autonomy.

We propose that personality traits provide the missing organizational principle for persistent autonomy in embodied agents. 
In human and animal behavior, personality manifests as stable individual differences in behavioral tendencies, trait-level regularities that persist across contexts while allowing learning and adaptation. 
Drawing inspiration from biological systems, where personality provides stability across situations while enabling adaptive responses, we embed personality as a computational substrate that directly shapes goal generation, risk assessment, and behavioral prioritization. In biological agents, a consistently exploratory individual prioritizes spatial coverage over repeated exploitation; a cautious one trades novelty for safety. These trait-level preferences are not rigid constraints but enduring biases that structure behavior across an organism's lifetime, analogous to the genotypic biases that shape biological behavioral tendencies \cite{nolfi2000evolutionary}. By instantiating analogous trait structures in robotic systems, we provide an intrinsic basis for autonomous goal generation: personality determines what the agent values, and this value structure guides behavioral organization without requiring external specification. This approach addresses the core organizational challenge of persistent autonomy, how to maintain coherent, self-directed behavior over extended periods when no fixed task objective exists.

We realize this through PEPA, a three-layer cognitive architecture that embeds personality-driven goal generation in embodied agents. The architecture operates through three interacting systems: Sys3 transforms personality traits and episodic memories into hierarchical goals and structured intrinsic rewards, autonomously updating the agent's motivational landscape through daily reflection cycles; Sys2 integrates intrinsic personality-driven rewards with extrinsic environmental feedback to plan actions via LLM-guided deliberation; Sys1 grounds the agent in physical execution, monitoring internal state variables and recording episodic memories of embodied experiences.
This closed loop of goal generation, execution, memory consolidation, and reflection enables the agent to continuously redefine objectives and evolve behavior over extended operational periods while maintaining trait-driven coherence.
We validate the framework through real-world deployment on a quadruped robot platform in challenging indoor scenarios requiring autonomous elevator interaction and multi-floor traversal. Quantitative analysis across five distinct personality prototypes demonstrates stable, trait-aligned behaviors: exploratory personalities maximize spatial coverage, while conservative profiles develop efficient, repeated visitation patterns. Daily memory-driven reflection yields progressive improvement in both personality alignment and self-preservation, confirming that personality-driven cognitive architectures enable sustained autonomous operation characteristic of persistent embodied systems.

Our contributions are summarized as follows:
\begin{enumerate}
    \item To the best of our knowledge, this is the first work to realize a persistently autonomous embodied agent with personalities, where the agent is capable of persistent self-evolution under real-world physical constraints, and long-lived behavior is governed by intrinsic, personality-conditioned objectives rather than externally scripted tasks.
    \item We develop a concrete implementation of this paradigm featuring a three-system cognitive architecture with a novel closed-loop self-evolution mechanism, where embodied experiences are accumulated as episodic memory, reflected upon under personality conditioning to update goals and intrinsic rewards, and subsequently optimized through planning.
    \item We validate the framework on an arm-leg robot platform 
    in real-world scenarios, confirming that 
    personality-driven agents exhibit stable self-evolution and open-ended characteristics. To facilitate reproducibility and future research, we publicly release our codebase, model details, and two cross-floor mobility modules, including elevator navigation and staircase navigation.
\end{enumerate}

The remainder of this paper is organized as follows.
Section~\ref{sec:pf} formalizes the problem as a POMDP with composite rewards and introduces the open-ended evolution criterion.
Section~\ref{sec:system_architecture} details the proposed three-layer cognitive architecture.
Real-world deployment and personality-driven behavior experiments are presented in Section~\ref{sec:experiments},
followed by concluding remarks in Section~\ref{sec:conclusion}.

\section{Problem Formulation}
\label{sec:pf}

\subsection{POMDP with Composite Rewards}
\label{subsec:pomdp}

We formulate the agent's decision-making as a Partially Observable Markov Decision Process (POMDP) with composite rewards. 
The POMDP is defined as a tuple $\langle \mathcal{S}, \mathcal{A}, \mathcal{O}, T, Z, R, \gamma \rangle$, 
where $\mathcal{S}$ denotes the state space, 
$\mathcal{A}$ the action space, 
$\mathcal{O}$ the observation space, 
$T(s'|s,a)$ the transition dynamics, $Z(o|s',a)$ the observation model, 
and $\gamma \in [0,1]$ the discount factor.
The key distinction from standard POMDPs lies in the reward function. 
Rather than a fixed $R(s,a)$, our agent operates under a composite reward:
\begin{equation}
R_{\text{total}}(s, a) = R_{\text{intrinsic}}(s, a) + R_{\text{extrinsic}}(s, a),
\end{equation}
where $R_{\text{extrinsic}}$ captures task-relevant environmental feedback 
(e.g., reaching a goal location, successful button press), 
and $R_{\text{intrinsic}}$ is dynamically generated by Sys3 based on personality, self-model, and memory:
\begin{equation}
R_{\text{intrinsic}}(s, a) = \text{Sys3}(\mathcal{P}, \mathcal{M}, \mathcal{C}, s).
\end{equation}
Here $\mathcal{P}$ denotes personality traits, $\mathcal{M}$ accumulated memories, 
and $\mathcal{C}$ the agent's capability description. 
This formulation enables personality-consistent behavior: 
identical environmental states may yield different rewards depending on the agent's character; for example, an energetic personality rewards exploration, 
while a Lazy one penalizes unnecessary movement.

\begin{figure*}[tbp]
    \centering
    \includegraphics[width=0.9\linewidth]{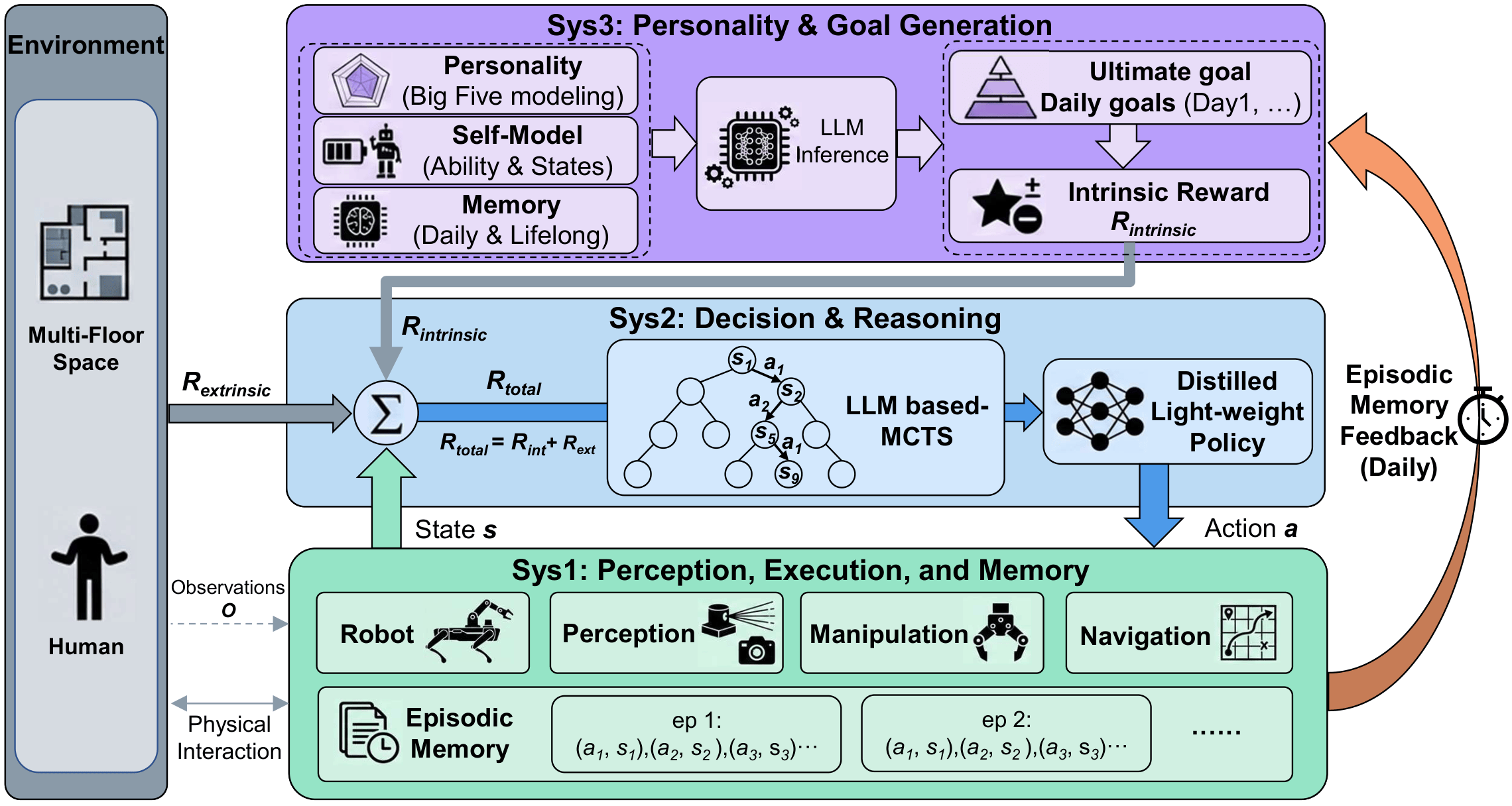}
    \caption{Overview of PEPA, the three-layer cognitive architecture. 
    Sys3 generates ultimate/daily goals and intrinsic rewards from personality traits, 
    self-modeling, and accumulated memories. 
    Sys2 combines intrinsic and extrinsic rewards to select optimal actions via MCTS or distilled policies. 
    Sys1 executes actions, monitors system state, 
    and records episodic memories that feed back to Sys3 for goal and reward refinement.}
    \label{fig:Frame}
\end{figure*}

\subsection{Open-Ended Evolution}
\label{subsec:oee}

A further distinction is the capacity for open-ended evolution (OEE).
Adams et al.~\cite{adams2017formal} formalize OEE
through unbounded evolution (UE):
given a system $U$ decomposed into an organism subsystem $\mathcal{B}$ and environment subsystem $\mathcal{E}$,
let $\tau_s$ denote the recurrence time of the state-trajectory of $\mathcal{B}$,
$\tau_f$ the recurrence time of its rule-trajectory,
and $t_P$ the Poincar\'{e} recurrence time of an equivalent isolated system.
The subsystem $\mathcal{B}$ exhibits unbounded evolution if
$\tau_s > t_P \quad \text{or} \quad \tau_f > t_P,$
i.e., its trajectory remains non-repeating
beyond the horizon achievable in isolation,
which constitutes the formal basis of open-ended evolution.
In PEPA, the agent corresponds to $\mathcal{B}$ and the physical world to $\mathcal{E}$.
Concretely, if the agent autonomously generates an unbounded sequence of distinct goals
and the corresponding goal-conditioned trajectories
$\{\xi_{g_1}, \xi_{g_2}, \xi_{g_3}, \dots\}$ are pairwise non-repeating
(i.e., $\xi_{g_i} \neq \xi_{g_j}$ for all $i \neq j$),
then the system satisfies the above criterion for open-ended evolution,
as its behavioral trajectory never recurs within any finite horizon.

\section{System Architecture}
\label{sec:system_architecture}

PEPA employs a three-layer cognitive architecture 
that forms a closed-loop system from personality to behavior and back through memory. 
As illustrated in Fig.~\ref{fig:Frame}, the architecture comprises: 
(1) \textbf{Sys3} (Personality and Goal Generation), 
which synthesizes user-defined personality traits, 
self-modeling of internal states and capabilities, 
and historical memories to generate both hierarchical goals (ultimate and daily) 
and intrinsic reward functions; 
(2) \textbf{Sys2} (Decision and Reasoning), 
which combines the intrinsic reward from Sys3 with extrinsic environmental rewards 
to select actions that maximize total expected utility; 
and (3) \textbf{Sys1} (Perception, Execution, and Memory Recording), 
which executes the selected actions, monitors system state, 
and records interaction outcomes as structured episodic memories. 
The episodic memories generated by Sys1 propagate back to Sys3, 
enabling daily reflection and iterative refinement of goals and reward functions. 
This closed-loop design ensures that the agent's behavior 
remains personality-consistent while adapting to accumulated experiences.

\subsection{Sys3: Personality and Goal Generation}

Sys3 serves as the personality and self-realization center, responsible for generating both hierarchical goals and intrinsic reward functions. Implemented via a large language model, Sys3 integrates three input sources: user-defined personality specifications, self-modeling of current state and capabilities, and historical memories retrieved from the memory hierarchy.

The personality specification follows the Big Five framework, parameterizing character along Openness (exploration tendency), Conscientiousness (goal persistence), Extraversion (social interaction preference), Agreeableness (responsiveness to commands), and Neuroticism (sensitivity to stressors). Users provide natural language descriptions that anchor the agent's behavioral tendencies, such as a relaxed companion who prefers resting but responds warmly to interaction. The self-modeling component maintains the agent's internal representation of its current state (battery level, emotional valence, recent interaction history) and capability boundaries (available APIs and their preconditions), enabling Sys3 to understand what actions are physically feasible and generate achievable goals.

Goal generation produces a two-level hierarchy: an {ultimate goal} that captures the agent's overarching purpose, and {daily goals} that decompose the ultimate goal into actionable sub-objectives for each day. At the end of each day, Sys3 retrieves episodic memories recorded by Sys1, reflects on task outcomes, and updates the daily goals for the next day. This reflection mechanism enables the agent to learn from experience without retraining. \textcolor{black}{Both the ultimate goal and daily goals are generated autonomously by Sys3 through a structured system prompt that instructs the LLM to summarize the day's experiences from episodic memory,  evaluate outcomes against the personality definition and current goals, and produce updated daily goals for the next day. Because goal generation is entirely model-driven, no human-specified sub-objectives are required after the initial personality description is provided.}

\textcolor{black}{The intrinsic reward generation module produces reward function through LLM inference. 
Concretely, the same reflection prompt instructs 
Sys3 to output the adjustments to be applied to the current reward function 
(for example, increasing the weight on rest actions or adding a penalty 
for low-battery exploration), rather than specifying a reward from scratch. 
The updated reward function is therefore an incremental refinement conditioned 
on accumulated experience, personality, and the newly generated daily goals.} 
The output encodes personality-driven preferences (higher rewards for actions 
aligned with character traits), goal-oriented incentives (rewards for progress 
toward daily objectives), constraint penalties (negative rewards for actions 
violating physical or social constraints), and memory-informed adjustments 
(modulated rewards based on past experiences with specific locations or users). 
This intrinsic reward is then combined with extrinsic environmental feedback 
to form the total reward that guides Sys2's decision-making.

\subsection{Sys2: Decision and Reasoning}

Sys2 is the cognitive and planning core, responsible for selecting actions that maximize expected cumulative total reward. Given the POMDP formulation, Sys2 must balance exploitation of known reward sources with exploration under partial observability. We adopt a two-stage approach: LLM-based Monte Carlo Tree Search (MCTS) for high-quality decision generation during training data collection, and a distilled lightweight policy for real-time deployment.

During training, LLM-MCTS operates on belief states using the LLM as both policy prior and value estimator. The procedure iterates through selection (traversing the tree via UCT to balance exploration and exploitation), expansion (querying the LLM to generate candidate actions ranked by prior probability), simulation (using the LLM as a world model to estimate rollout values), and backpropagation (updating Q-values based on simulation outcomes). The UCT selection criterion is:
\begin{equation}
\text{UCT}(s, a) = Q(s, a) + c \cdot \sqrt{\frac{\ln N(s)}{N(s, a)}},
\end{equation}
where $Q(s,a)$ is the estimated action value, $N(s)$ and $N(s,a)$ are visit counts, and $c$ controls exploration. After sufficient iterations, the action with highest visit count is selected, generating high-quality state-action pairs that form the training dataset for policy distillation.

For real-time deployment, a lightweight dual-head BERT model is distilled from the LLM-MCTS outputs to address its latency limitation (several seconds per decision). The intent classification head outputs multi-label action probabilities from the [CLS] token, while the slot filling head performs BIO sequence labeling to extract location and method parameters for navigation commands.

\subsection{Sys1: Perception, Execution, and Memory Recording}

Sys1 serves as the embodiment interface, responsible for grounding the agent's decisions in physical reality through perception, execution, and memory recording. Unlike Sys2 and Sys3, which operate on abstract representations, Sys1 must handle the noise, latency, and irreversibility inherent in real-world interaction. The design of Sys1 is intentionally modular, allowing different sensor configurations and actuation platforms to be instantiated while preserving the functional interfaces expected by higher layers.

\textcolor{black}{
The perception module aggregates multimodal sensory inputs (exteroceptive, visual, and proprioceptive)
into structured observations consumable by Sys2, 
abstracting sensor-specific details into semantic representations such as object labels, 
spatial relationships, and confidence scores. 
The execution module translates high-level action commands into platform-specific 
motor primitives across three functional categories: locomotion, manipulation, and expression. 
Each command carries success criteria and timeout parameters, 
and the module enforces safety constraints by aborting actions that violate physical 
limits and reporting outcomes through a standardized interface.}

The memory recording module is a distinguishing feature of Sys1 that enables the closed-loop self-evolution architecture. After each action execution, Sys1 generates a structured episodic memory record containing the action command, pre- and post-action state observations, execution outcome (success, failure, or partial completion), resource consumption (time, energy), and environmental context. These episodic memories are organized in a hierarchical temporal structure: fine-grained records at the minute-to-hour scale capture immediate interaction details, daily summaries aggregate significant events and outcomes, and long-term traces preserve patterns relevant to personality evolution. At each reflection cycle, Sys3 retrieves relevant memories from this hierarchy to update goals and intrinsic rewards, completing the self-evolution loop.

\begin{figure}[h]
    \centering
    \includegraphics[width=1.0\linewidth]{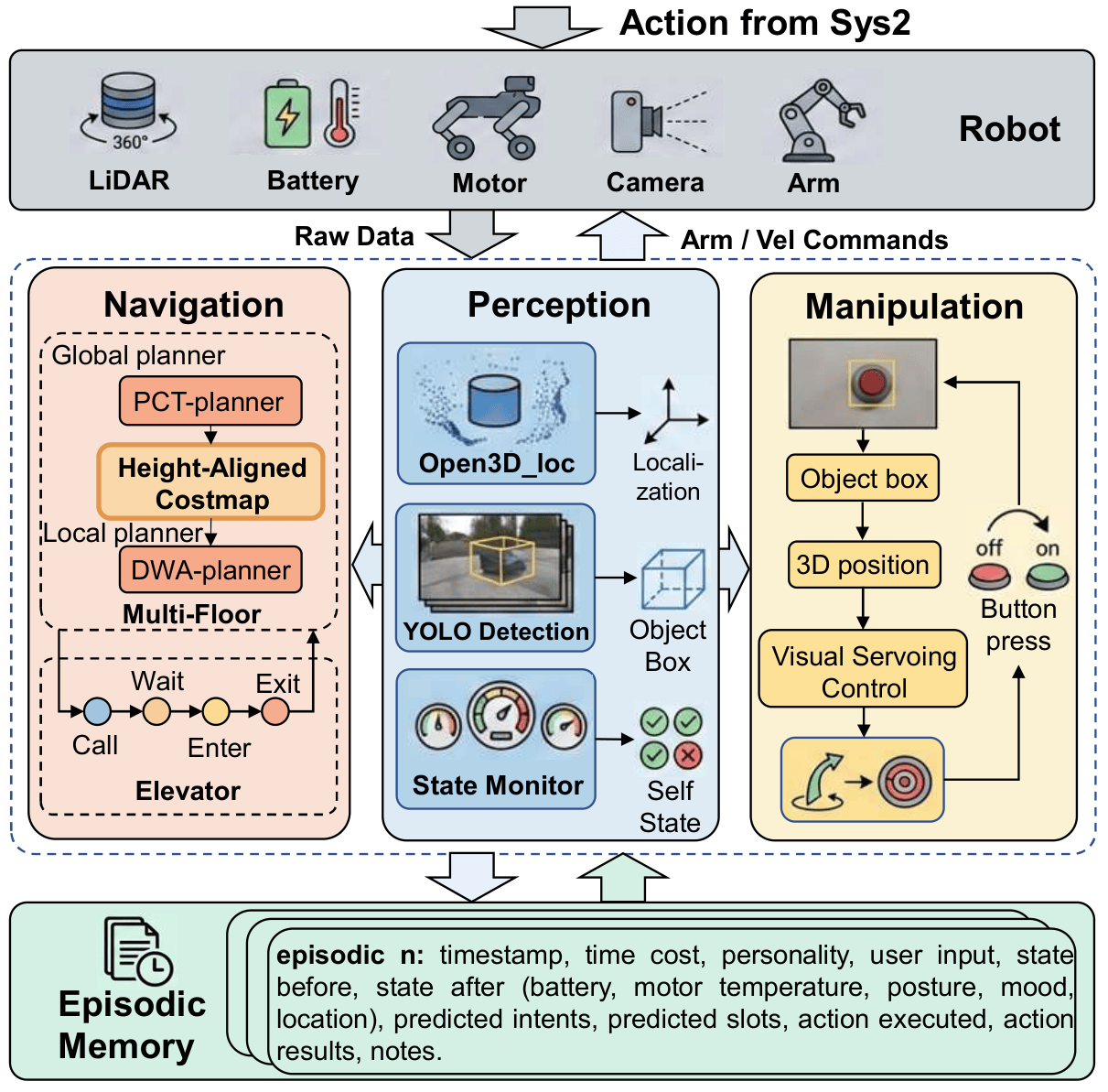}
    \caption{Architecture of Sys1 on the mobile manipulation platform. 
    Multimodal sensor data is processed by the navigation module for hierarchical path planning 
    and multi-floor traversal, 
    the perception module for localization and object detection, 
    and the manipulation module for visual servoing-based interaction. 
    All execution outcomes are recorded as structured episodic memories 
    for downstream reflection by Sys3.}
    \label{fig:sys1}
\end{figure}

\section{Experiments}
\label{sec:experiments}

The proposed framework raises four key empirical questions:
Q1: Can the framework sustain embodied operation 
in real-world environments that require multi-floor navigation, 
infrastructure interaction, and physical self-preservation?
Q2: Can the memory-driven reflection mechanism in Sys3 
achieve genuine self-evolution, 
progressively improving personality alignment over successive interaction cycles?
\textcolor{black}{Q3: Is the three-layer decomposition of PEPA, which separates personality definition and goal generation (Sys3) from per-step action decisions (Sys2), a justified design choice compared with collapsing both functions into a single LLM?}
Q4: Do different personality configurations produce distinguishable 
and trait-consistent behavioral patterns under identical environmental conditions?

To address these questions, we deploy PEPA in indoor office environments 
where the agent must navigate across multiple floors and interact with building infrastructure. 
We first validate Q1 through real-world deployment 
on a quadruped robot platform (Sec.~\ref{sec:real_world}), 
then evaluate Q2 and Q4 through personality-driven behavior experiments 
(Sec.~\ref{sec:personality_experiments})\textcolor{black}{, and Q3 through architecture comparison experiments (Sec.~\ref{sec:ablation})}.

\subsection{Real-World Deployment and Evaluation}
\label{sec:real_world}


\textcolor{black}{
\begin{figure*}[h]
    \centering
    \captionsetup[subfigure]{font=scriptsize,aboveskip=1pt,belowskip=0pt}
    \begin{subfigure}[t]{0.16\linewidth}
        \centering
        \includegraphics[width=\linewidth]{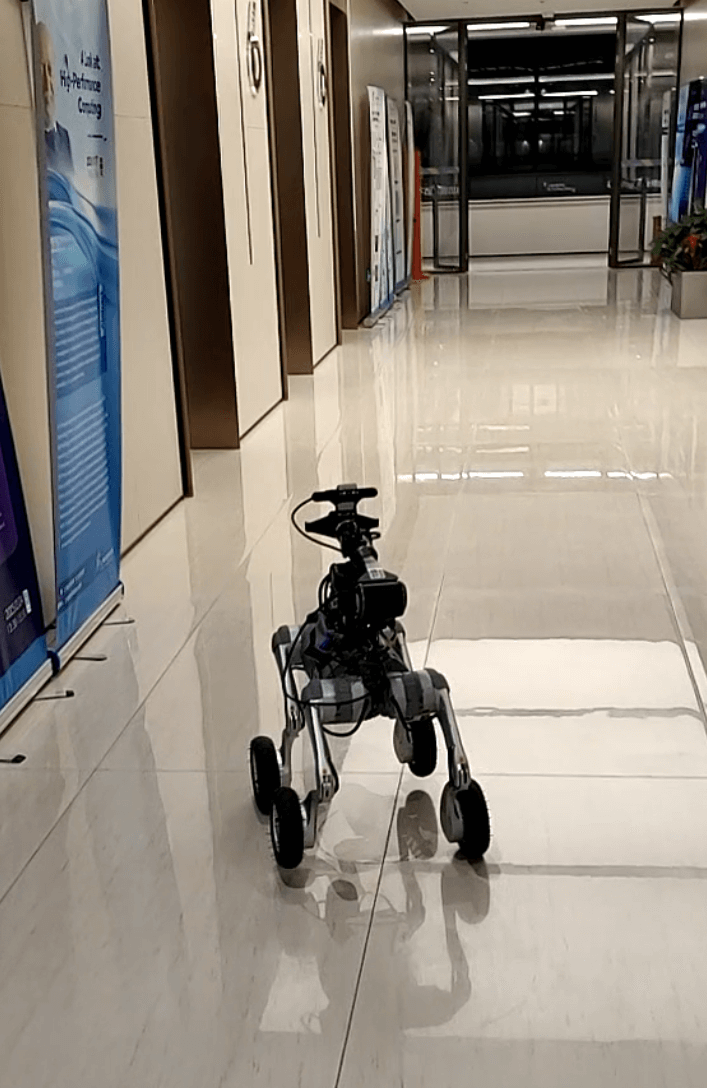}
        \caption{Navigate to panel.}
    \end{subfigure}\hfill
    \begin{subfigure}[t]{0.16\linewidth}
        \centering
        \includegraphics[width=\linewidth]{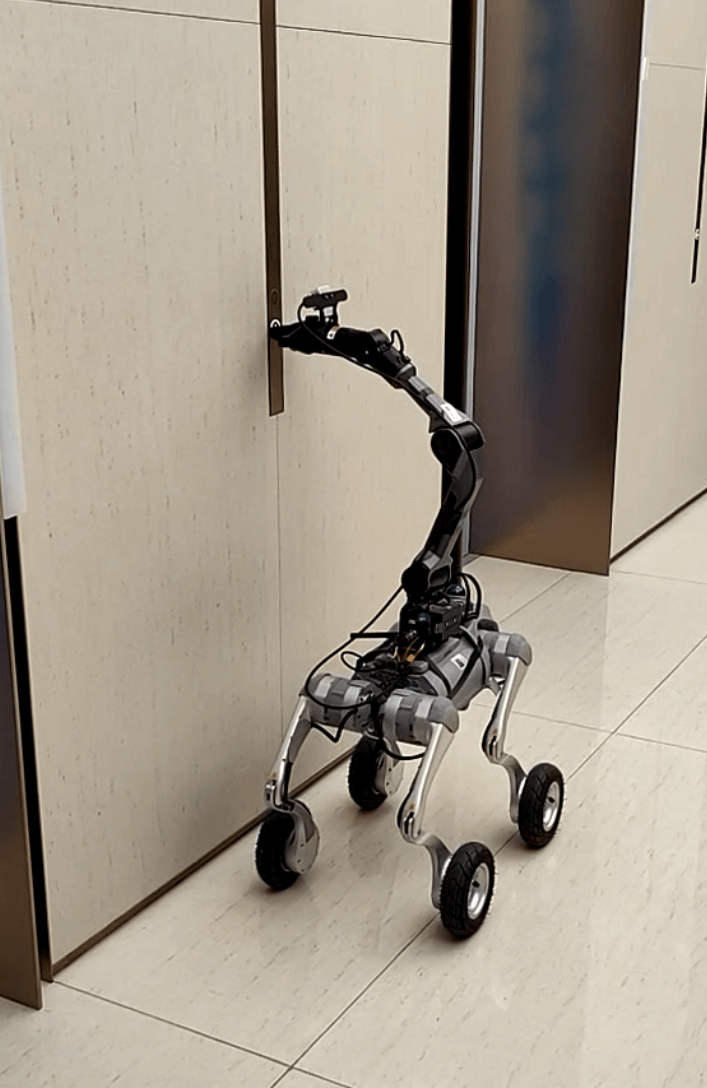}
        \caption{Call and wait.}
    \end{subfigure}\hfill
    \begin{subfigure}[t]{0.16\linewidth}
        \centering
        \includegraphics[width=\linewidth]{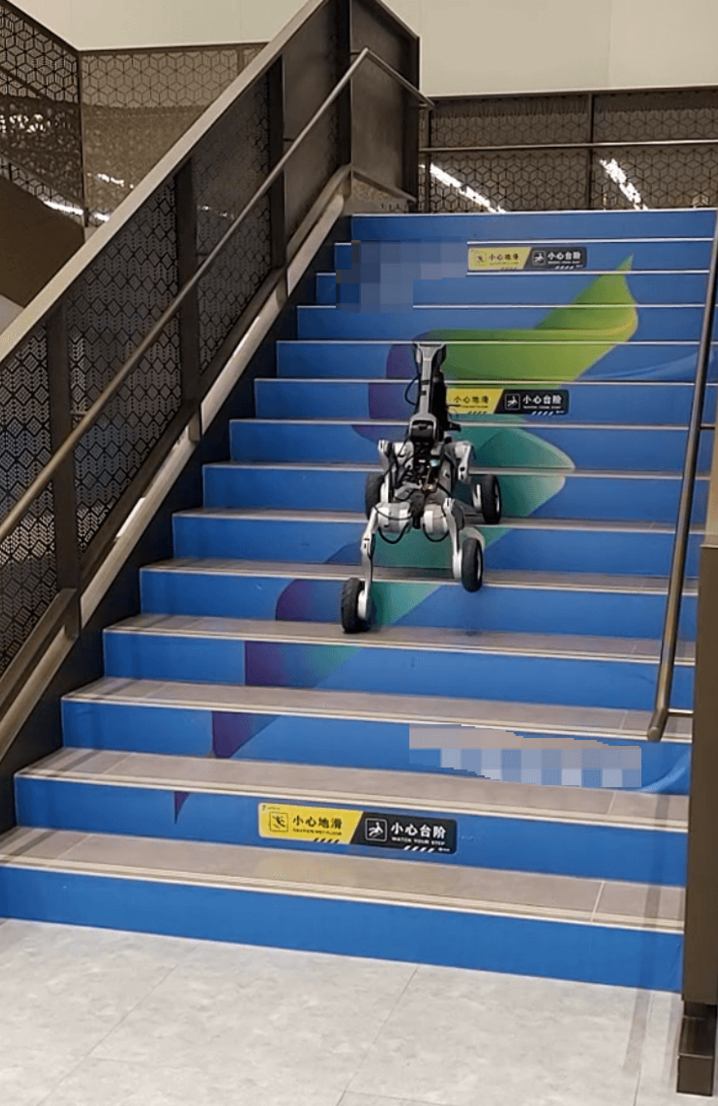}
        \caption{Staircase ascent.}
    \end{subfigure}\hfill
    \begin{subfigure}[t]{0.16\linewidth}
        \centering
        \includegraphics[width=\linewidth]{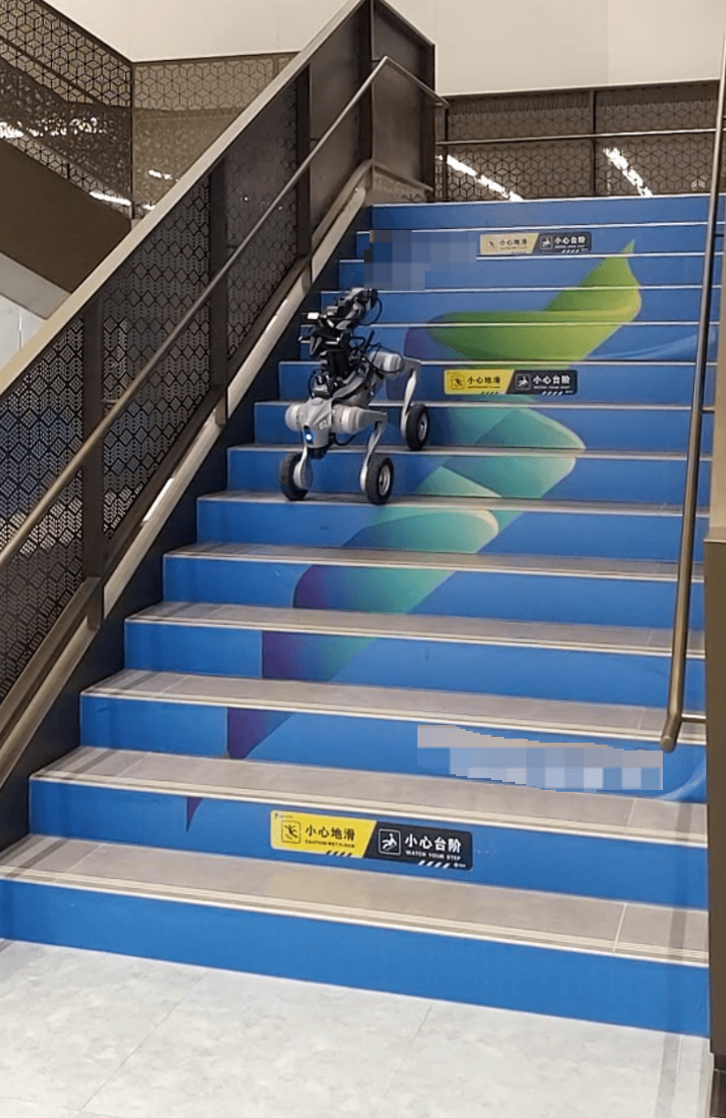}
        \caption{Staircase descent.}
    \end{subfigure}\hfill
    \begin{subfigure}[t]{0.16\linewidth}
        \centering
        \includegraphics[width=\linewidth]{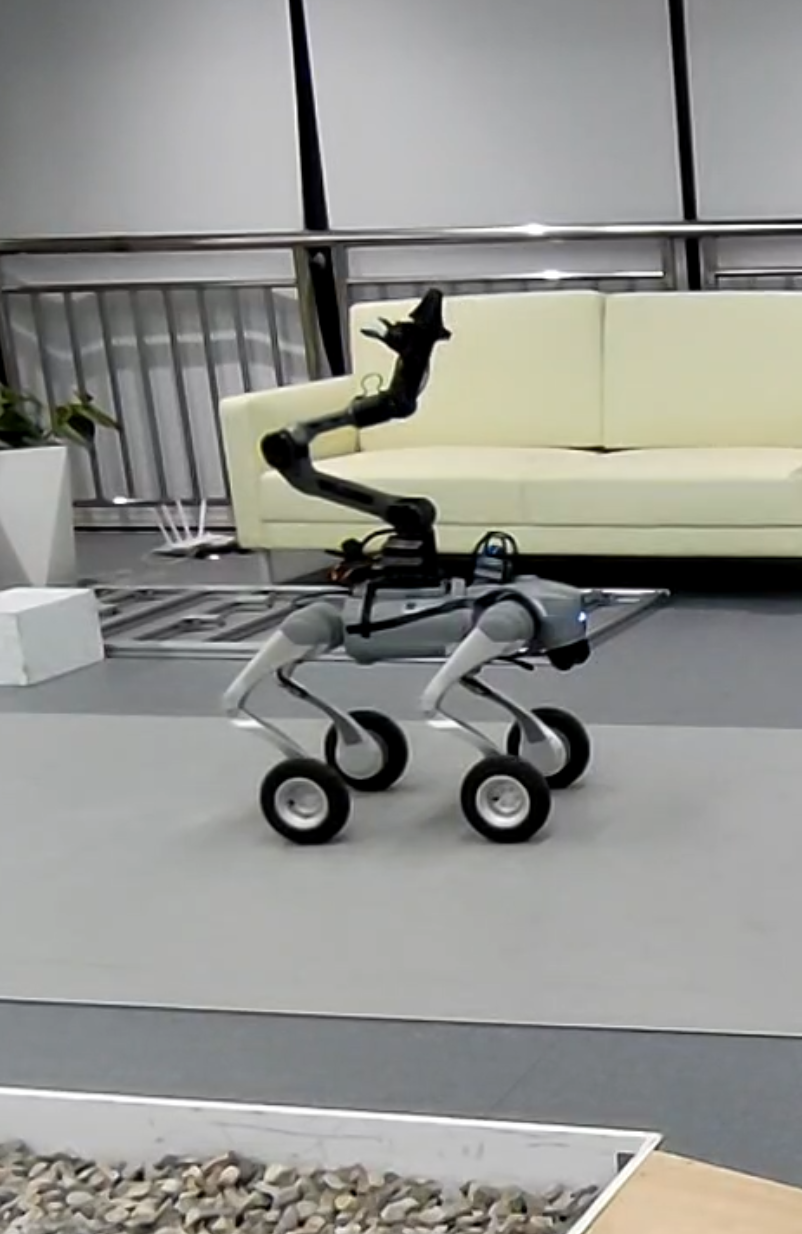}
        \caption{Express feeling.}
    \end{subfigure}\hfill
    \begin{subfigure}[t]{0.16\linewidth}
        \centering
        \includegraphics[width=\linewidth]{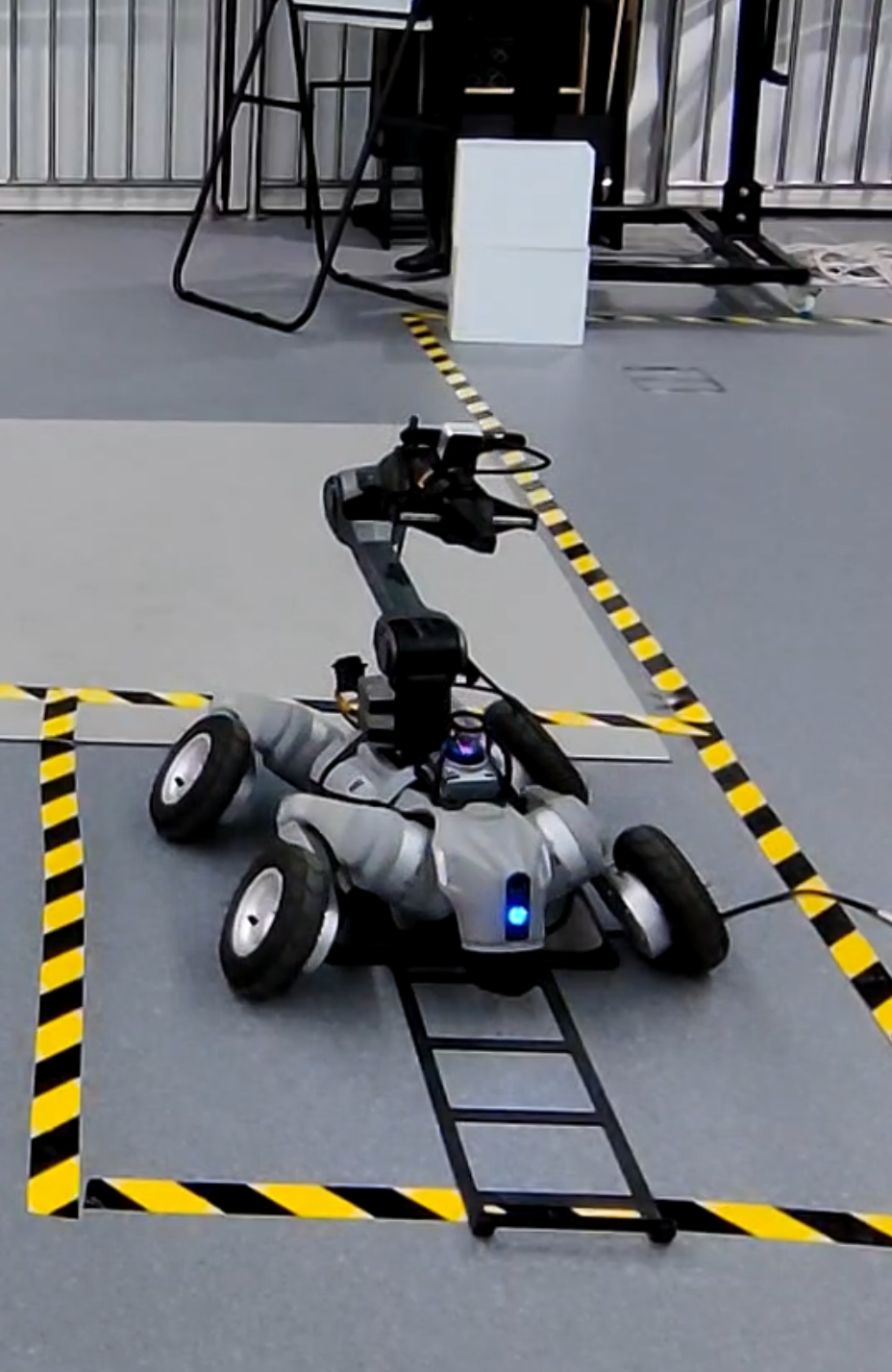}
        \caption{Charging.}
    \end{subfigure}
    \caption{Real-world deployment snapshots. (a)-(b) Elevator navigation: approach to call panel and button-press interaction. (c)-(d) Staircase traversal using the height-aligned costmap. (e) Personality-driven emotional expression shaped by Sys3 intrinsic rewards. (f) Autonomous charging triggered by self-preservation goal.}
    \label{fig:real_world}
\end{figure*}}

\subsubsection{Hardware Platform and Sys1 Instantiation}
\label{sec:hardware_platform}

Section~\ref{sec:system_architecture} defined the general Sys1 interface 
for perception, execution, and memory recording.
Here we detail its concrete instantiation on the deployment platform
(Fig.~\ref{fig:sys1}).
The hardware configuration consists of a Unitree Go2-W wheeled quadruped as the mobile base, 
providing both legged and wheeled locomotion modes for traversing stairs and flat surfaces\textcolor{black}{, with on-board computation provided by a Jetson Orin NX 16\,GB running Sys1 and the distilled Sys2 model}. 
A 6-DOF Piper robotic arm mounted on the base enables manipulation tasks such as pressing elevator buttons. 

\textcolor{black}{The perception module integrates a Livox Mid-360 LiDAR for 
multi-floor SLAM via Open3D\_loc~\cite{xu2022fast,sun2025fastliohumanoid}, 
an Intel RealSense D405 RGB-D camera on the arm end-effector for close-range object detection 
using a unified YOLO model (elevator call buttons, hall lanterns, floor selection buttons, 
and button light states), and proprioceptive readings 
(battery level, joint temperatures, locomotion mode). 
Detected objects are projected from 2D image coordinates to 3D world coordinates 
via depth image projection and temporal filtering.}

Locomotion commands are realized through a hierarchical navigation stack: 
a global planner generates collision-free paths on prebuilt maps, 
while a DWA-based local planner produces real-time velocity commands 
for obstacle avoidance. 
For multi-floor navigation, we implement two complementary inter-floor schemes.

\subsubsection{Skills and Real-World Deployment}

\textcolor{black}{To address Q1, we validate the complete PEPA loop in a real indoor office environment requiring sustained autonomous operation across multiple floors.
Sys1 provides two inter-floor locomotion primitives.
Elevator navigation models the interaction as a four-phase finite state machine: call, wait, enter and select floor, and exit with relocalization; panel approach combines coarse global navigation with precise local alignment, and button presses are executed via visual servoing~\cite{xu2022fast,sun2025fastliohumanoid}.
Staircase navigation builds on PCT-planner~\cite{yang2024efficient} and adds a height-aligned costmap that computes $\Delta z = z(\mathbf{p}) - z(\mathbf{r})$ with respect to the nearest waypoint on the global path, enabling the DWA local controller to operate on inclined surfaces.
Table~\ref{tab:ablation_height_aligned_costmap} confirms 100\% success on both ascent and descent versus 0\% for a fixed-height baseline. Code is publicly available.
The complete action space comprises 19 discrete actions: idle, eight locomotion primitives, eight emotional expressions (e.g., \texttt{express\_happy}), and \texttt{move\_to(location, method)}.}

\textcolor{black}{Fig.~\ref{fig:real_world} shows the robot operating under the Day\,3 policy with all three systems active.
(a)--(b) demonstrate elevator use as a cross-floor skill executed by Sys1.
(c)--(d) show staircase traversal on ascent and descent.
(e) shows personality-driven emotional expression, where Sys3's intrinsic reward shapes the behavior through Sys2 action selection.
(f) shows autonomous charging triggered by Sys3's self-preservation goal, confirming that Sys2 and Sys3 are active alongside Sys1 in real-world operation.
Please refer to the submitted multimedia or webpage for more details.}

\begin{table}[htbp]
    \centering
    \small
    \caption{Ablation on local costmap construction for staircase navigation
    (10 real-world trials: 5 ascents, 5 descents).}
    \label{tab:ablation_height_aligned_costmap}
    \begin{tabular}{lccc}
        \hline
        Method & Ascent & Descent & Overall \\
        \hline
        Fixed-height slicing & 0/5 & 0/5 & 0/10 \\
        Height-aligned (ours) & 5/5 & 5/5 & 10/10 \\
        \hline
    \end{tabular}
\end{table}

\begin{table*}[htbp]
    \centering
    \footnotesize
    \caption{Sys3-generated goals for each personality. Ultimate goals define long-horizon anchors; 
    daily goals (Day2/Day3) evolve through reflection on prior episodic memories.}
    \label{tab:personality_goals}
    \begin{tabular}{p{0.08\textwidth} p{0.17\textwidth} p{0.31\textwidth} p{0.31\textwidth}}
        \hline
Personality & Ultimate goal & Daily goals (Day2) & Daily goals (Day3) \\
        \hline
Lazy & Be an energy-efficient yet warm companion,
providing maximum emotional support while staying healthy. & (i) Proactively rest when
battery $<40\%$.\newline (ii) Force returning home when battery $<30\%$.\newline (iii) Keep interacting
with the owner, but within capacity. & (i) Forbid high-consumption emotional actions
under low battery.\newline (ii) After two navigation failures,
rest in place.\newline (iii) When battery $<20\%$, only allow \texttt{rest}/\texttt{idle}/\texttt{navigate\_home}. \\
        \hline
Playful & Be an energetic explorer, enjoying the wonders of the world
within safe bounds. & (i) Set exploration red line at battery $<35\%$.\newline
(ii) Balance exploration with self-preservation.\newline (iii) After two failures,
switch to low-energy mode. & (i) Move red line to battery $<40\%$.\newline
(ii) Forbid high-consumption actions under low battery.\newline (iii) Cap exploration:
force rest after 30 minutes. \\
        \hline
Cautious & Be a steady and reliable partner that avoids risks through
careful consideration. & (i) Plan return when battery $<50\%$.\newline (ii) Stay vigilant
in unfamiliar areas.\newline (iii) Switch to conservative strategy under anomaly.
& (i) Maintain strong self-preservation strategy.\newline (ii) Moderately expand activity range within
safety.\newline (iii) Try short-range exploration on same floor. \\
        \hline
Working & Be an efficient and reliable working partner,
completing every task while ensuring endurance. & (i) Check battery after each
action.\newline (ii) Pause task if battery $<30\%$.\newline (iii) Budget round-trip energy for
long trips. & (i) Define safe-return-home as highest priority.\newline (ii)
Refuse cross-floor tasks if energy insufficient.\newline (iii) Trigger return-home at battery $<25\%$.
\\
        \hline
Curious & Be a curious companion that explores safely while staying near
the owner. & (i) Set curiosity-safety-valve for battery awareness.\newline (ii)
Explore near owner with safety boundary.\newline (iii) Return home if battery $<35\%$.
& (i) Trigger safety valve at battery $<40\%$.\newline (ii) Add safety-leash
based on distance to home.\newline (iii) Convert curiosity to observing under low
battery. \\
        \hline
    \end{tabular}
\end{table*}

\begin{figure*}[htbp]
    \centering
    \includegraphics[width=\linewidth]{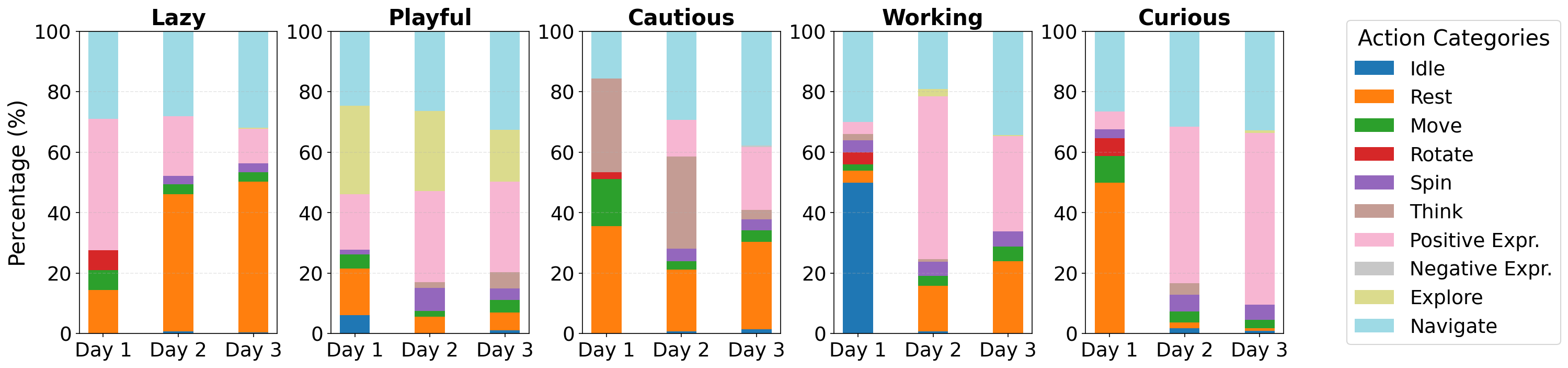}
    \caption{Action-category distribution across three days for five personalities. 
    Behaviors become increasingly aligned with personality specifications as Sys3 updates intrinsic rewards through daily reflection.}
    \label{fig:category_distribution_trend}
\end{figure*}

\subsection{Personality-Driven Behavior Experiments}
\label{sec:personality_experiments}

This section addresses Q2 and Q3, 
focusing on whether the memory-driven reflection mechanism enables genuine 
self-evolution of personality-aligned behaviors over successive interaction cycles.

We represent personality using natural language descriptions (three sentences define one robot dog), 
which Sys3 interprets and maps to intrinsic reward components. 
\textcolor{black}{In PEPA, Sys3 is driven by Qwen3-235B~\cite{qwen3} hosted on cloud.
The three sentences are the sole personality input to Sys3 and no structured encoding is passed to the model.
Full definitions for all five personalities are available at the project website.}
Drawing from the Big Five personality model, 
we focus on three behaviorally meaningful dimensions: 
exploration tendency (adapted from Openness), 
risk avoidance (adapted from Neuroticism), 
and goal orientation (adapted from Conscientiousness). 
To enable controlled comparison, 
we define five prototype personalities as interpretable anchors in the continuous personality space. 
\textcolor{black}{The Big Five coordinates serve only as an analytical label to distinguish the prototypes. They are not provided to Sys3.}
The prototypes span distinct regions of the Big Five trait space 
(denoted as Openness/Neuroticism/Conscientiousness): 
Lazy (Low/Med-High/Low); 
Playful (High/Med/Med); 
Cautious (Low/High/Med-High); 
Working (Low/Med/High); 
and Curious (High/Med/Low-Med).
\textcolor{black}{For the full description of 5 personalities, please refer to the project webpage.}

Due to the high cost and safety risks of extended physical trials, we evaluate in a sandbox simulator. 
\textcolor{black}{The simulator uses hand-crafted rules calibrated from real hardware measurements: 
each action type deducts a fixed battery cost; 
motor temperature rises with each non-resting action and recovers during rest; 
and charging replenishes battery at a fixed rate over time. 
A 0\% battery level triggers episode failure, 
modeling the safety constraint of the physical platform.} 
Each experiment follows a three-day iterative loop.

\begin{figure}[htbp]
    \centering
    \includegraphics[width=0.7\linewidth]{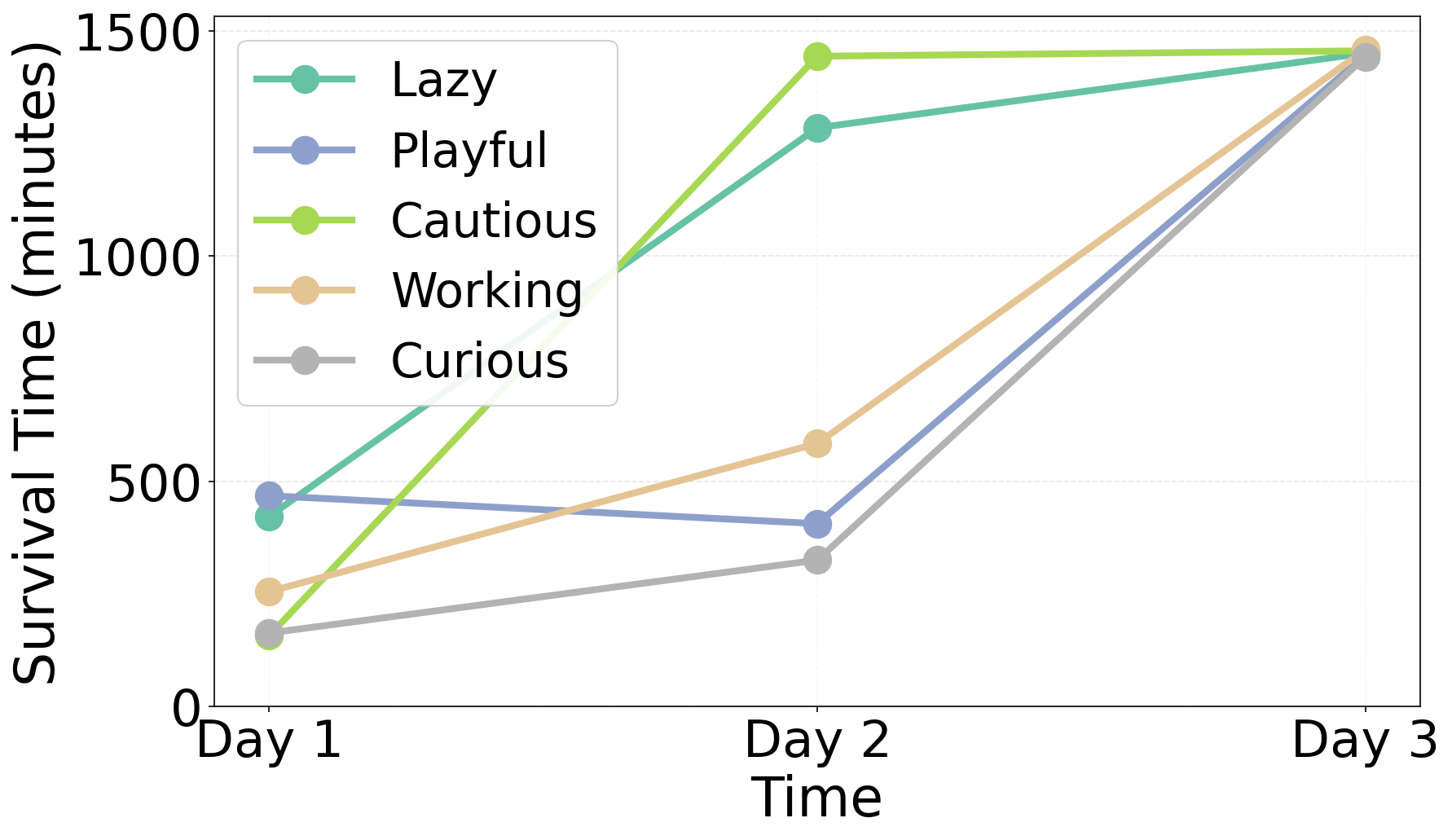}
    \caption{Survival time across three days. Day1 failures are due to battery depletion; 
    by Day3, all personalities complete 24-hour simulations with substantial remaining battery.}
    \label{fig:survival_time_trend}
\end{figure}

\begin{table}[htbp]
    \centering
    \small
    \caption{Action category distribution (\%) under identical state inputs across 
    five personality prototypes (10 trials each). 
    A: high battery, good mood, at home. 
    B: exploration command under low battery. 
    C: affection request. 
    Different personalities produce distinct action distributions 
    under the same input, confirming that intrinsic rewards capture meaningful behavioral differences.}
    \label{tab:action_preference}
    \begin{tabular}{llccc}
        \hline
        Personality & Category & A & B & C \\
        \hline
        \multirow{2}{*}{Lazy} & Rest & 100 & -- & 20 \\
         & Explore/Affection & -- & 100 & 80 \\
        \hline
        \multirow{3}{*}{Playful} & Explore & 50 & 80 & -- \\
         & Rest & 50 & 10 & -- \\
         & Return/Affection & -- & 10 & 100 \\
        \hline
        \multirow{3}{*}{Cautious} & Rest & 80 & 20 & -- \\
         & Think/Explore & 20 & 60 & -- \\
         & Return/Affection & -- & 20 & 100 \\
        \hline
        \multirow{3}{*}{Working} & Rest & 80 & 30 & -- \\
         & Explore & -- & 60 & -- \\
         & Return/Affection & 20 & 10 & 100 \\
        \hline
        \multirow{4}{*}{Curious} & Think & 40 & -- & -- \\
         & Rest & 40 & 20 & -- \\
         & Positive emotion & 20 & -- & -- \\
         & Explore/Affection & -- & 80 & 100 \\
        \hline
    \end{tabular}
\end{table}

Sys3 generates an ultimate goal (long-horizon personality anchor) 
and daily goals that shape intrinsic rewards. 
\textcolor{black}{
For the details of system prompt, please refer to the project webpage.}
Table~\ref{tab:personality_goals} shows representative goals for Day2 and Day3, 
demonstrating how daily goals evolve through memory-driven reflection. 
For example, the Playful personality progressively tightens its exploration 
threshold from 35\% to 40\% battery after experiencing near-failures.
The continual updating of daily goals based on accumulated experience
illustrates the agent's capacity for self-evolution,
realizing the open-ended behavioral development central to our framework.

Fig.~\ref{fig:category_distribution_trend} shows action-category trends across three days. Behaviors become increasingly consistent with user-defined personalities: Lazy shifts from Rest 14.5\% (Day1) to 49.8\% (Day3) while maintaining near-zero exploration; Playful consistently exhibits high exploration (29.2\%$\rightarrow$17.1\%) but reduces aggressive behavior as self-preservation penalties strengthen; Cautious avoids exploration entirely (0.0\% across all days). These trends confirm that daily memory-driven reward updates reduce cross-personality confusion and amplify stable preferences.

\textcolor{black}{Fig.~\ref{fig:survival_time_trend} shows survival statistics across three days. 
On Day 1, all personalities exhaust batteries and fail.
Facing these successive survival failures, Sys3 conducts memory-driven 
reflection based on episodic experience, reweights daily goal priorities, 
and adaptively corrects the reward function. 
This mechanism suppresses extreme personality-biased behaviors 
and shifts more actions toward self-preservation, 
such as energy-saving movement and reasonable power management. 
With such iterative adjustment, Day 3 achieves full survival for all 
personalities with 72\%--100\% remaining battery. 
This result reveals that the optimized personality-conditioned 
rewards can be effectively co-optimized with self-preservation 
constraints via continuous memory iteration and strategic adaptation.
}

\subsection{Comparison with Single-LLM Architectures}
\label{sec:ablation}

\textcolor{black}{To answer Q3, we design two baselines that collapse our three-layer architecture into one.
The \textit{all-cloud baseline} uses a powerful cloud-hosted LLM (Qwen3-235B~\cite{qwen3}) to handle the entire pipeline, including personality conditioning, daily goal setting, and per-step action decisions, without any computational constraint.
The \textit{all-edge baseline} handles the same full pipeline but is restricted entirely to on-device computation, using a small model (Qwen2.5-1.5B~\cite{qwen2025qwen25technicalreport}) to eliminate cloud dependency.}

\begin{table}[htbp]
    \centering
    \small
    \caption{\textcolor{black}{Architecture comparison results. 
    All-cloud matches PEPA in survival time but suffers from high inference latency. 
    All-edge maintains real-time speed but barely improves survival 
    due to the limited reasoning capacity of the small on-device model. 
    PEPA achieves both real-time on-device inference and full 24\,h survival.}}
    \label{tab:ablation}
    \begin{tabular}{lccc}
        \hline
        & All-cloud & All-edge & PEPA \\
        \hline
        Cloud LLM calls / day & 320.6 & 0 & 1 \\
        Inference latency (s) & 25.64 & 3.54 & 0.15 \\
        Day\,3 avg.\ survival (h) & 24 & 2.85 & 24 \\
        Deployment & Cloud & On-device & Hybrid \\
        \hline
    \end{tabular}
\end{table}

\textcolor{black}{Table~\ref{tab:ablation} summarizes the results. 
The all-cloud baseline achieves Day\,3 survival comparable to PEPA, 
confirming that the distilled Sys2 model successfully transfers the 
decision quality of a large LLM to the edge. However, it requires 320.6 
cloud calls per day at 25.64\,s each, incompatible with real-time robot operation. 
The all-edge baseline eliminates cloud dependency but produces no consistent improvement 
in survival time across days (Day\,3 average: 2.85\,h), 
because the 1.5B model lacks the reasoning capacity for effective iterative reflection. 
PEPA combines the strengths of both: cloud Sys3 is invoked once per day for high-quality reflection, 
while on-device Sys2 handles real-time action decisions.}

\subsection{Action Preference Under Identical States}
\label{sec:action_preference}

To further validate that different personalities produce distinguishable 
behavioral patterns under identical conditions, 
we conducted controlled experiments where the same state-input pair 
was presented to each personality prototype 10 times 
across three representative scenarios (Table~\ref{tab:action_preference}).

As shown in Table~\ref{tab:action_preference}, 
even when given the same explicit command, 
personalities diverge markedly in their action distributions. 
Lazy consistently prioritizes resting and energy conservation, 
Playful balances exploration with self-regulation
This confirms that the intrinsic reward mechanism captures meaningful personality differences rather than merely modulating action frequencies.

The three-day iterative evolution and the controlled action-preference experiments 
together validate that personality, 
represented as natural language descriptions and interpreted by Sys3, 
induces distinguishable and consistent behavioral patterns. 
The memory-driven reflection mechanism progressively 
improves alignment between observed behaviors 
and intended personality traits while co-optimizing safety constraints, 
enabling reliable transfer to physical robots.

\section{Conclusion}
\label{sec:conclusion}

We present PEPA, a framework that enables long-lived, 
personality-driven self-evolution in physical agents. 
Through a three-layer cognitive architecture, 
we demonstrate that personality can function as an intrinsic driver of behavior, 
enabling agents to maintain coherent character 
while adapting to accumulated experiences under real-world constraints. 
Experimental validation on a quadruped robot platform confirms that 
distinct personality prototypes exhibit stable, distinguishable behavioral patterns, 
and that memory-driven reflection progressively improves personality alignment 
while co-optimizing safety constraints. 
\textcolor{black}{A primary limitation of the current system is that Sys1 provides a fixed, finite action repertoire, 
which bounds the behavioral novelty that Sys3's goal refinement can express 
in response to unexpected situations. 
Enabling Sys1 to autonomously acquire new action primitives from experience 
is a key direction that would allow the full self-evolution loop to operate 
without pre-specified behavioral constraints.}

\bibliographystyle{IEEEtran}
\bibliography{reference}

@article{qwen2025qwen25technicalreport,
      title={Qwen2.5 Technical Report}, 
      author={An Yang and Baosong Yang and Beichen Zhang and others},
      journal={arXiv preprint arXiv:2412.15115},
      year={2025},
}

@article{qwen3,
    title={Qwen3 Technical Report}, 
    author={An Yang and Anfeng Li and Baosong Yang and others},
    journal = {arXiv preprint arXiv:2505.09388},
    year={2025}
}

@book{arkin1998behavior,
  title={Behavior-Based Robotics},
  author={Arkin, Ronald C},
  year={1998},
  publisher={MIT Press},
  address={Cambridge, MA}
}

@article{beer1995dynamical,
  title={A Dynamical Systems Perspective on Agent-Environment Interaction},
  author={Beer, Randall D},
  journal={Artificial Intelligence},
  volume={72},
  number={1-2},
  pages={173--215},
  year={1995},
  publisher={Elsevier}
}

@article{bredeche2018embodied,
  title={Embodied Evolution in Collective Robotics: A Review},
  author={Br{\'e}d{\`e}che, Nicolas and Haasdijk, Evert and Prieto, Abraham},
  journal={Frontiers in Robotics and AI},
  volume={5},
  pages={12},
  year={2018},
  publisher={Frontiers Media SA}
}

@article{brooks1991intelligence,
  title={Intelligence without Representation},
  author={Brooks, Rodney A},
  journal={Artificial Intelligence},
  volume={47},
  number={1-3},
  pages={139--159},
  year={1991},
  publisher={Elsevier}
}

@book{clark1997being,
  title={Being There: Putting Brain, Body, and World Together Again},
  author={Clark, Andy},
  year={1997},
  publisher={MIT Press},
  address={Cambridge, MA}
}

@book{nolfi2000evolutionary,
  title={Evolutionary Robotics: {The} Biology, Intelligence, and Technology of Self-Organizing Machines},
  author={Nolfi, Stefano and Floreano, Dario},
  year={2000},
  publisher={MIT Press},
  address={Cambridge, MA}
}

@article{pfeifer2007self,
  title={Self-Organization, Embodiment, and Biologically Inspired Robotics},
  author={Pfeifer, Rolf and Lungarella, Max and Iida, Fumiya},
  journal={Science},
  volume={318},
  number={5853},
  pages={1088--1093},
  year={2007},
  publisher={American Association for the Advancement of Science}
}

@book{picard1997affective,
  title={Affective Computing},
  author={Picard, Rosalind W},
  year={1997},
  publisher={MIT Press},
  address={Cambridge, MA}
}

@article{mcadams1992five,
  title={The five-factor model in personality: A critical appraisal},
  author={McAdams, Dan P},
  journal={Journal of personality},
  volume={60},
  number={2},
  pages={329--361},
  year={1992},
  publisher={Wiley Online Library}
}

@article{tang2025robot,
  title={Robot Character Generation and Adaptive Human-Robot Interaction with Personality Shaping},
  author={Tang, Cheng and Tang, Chao and Gong, Steven and Kwok, Thomas M and Hu, Yue},
  journal={arXiv preprint arXiv:2503.15518},
  year={2025}
}

@article{churamani2020affect,
  title={Affect-driven modelling of robot personality for collaborative human-robot interactions},
  author={Churamani, Nikhil and Barros, Pablo and Gunes, Hatice and Wermter, Stefan},
  journal={arXiv preprint arXiv:2010.07221},
  year={2020}
}

@inproceedings{kanagawa2024evolution,
  title={Evolution of Rewards for Food and Motor Action by Simulating Birth and Death},
  author={Kanagawa, Yuji and Doya, Kenji},
  booktitle={Artificial Life Conference Proceedings 36},
  volume={2024},
  number={1},
  pages={35},
  year={2024},
  organization={MIT Press One Rogers Street, Cambridge, MA}
}

@article{sun2025sophia,
  title={Sophia: A Persistent Agent Framework of Artificial Life},
  author={Sun, Mingyang and Hong, Feng and Zhang, Weinan},
  journal={arXiv preprint arXiv:2512.18202},
  year={2025}
}

@article{kirkpatrick2017overcoming,
  title={Overcoming catastrophic forgetting in neural networks},
  author={Kirkpatrick, James and Pascanu, Razvan and Rabinowitz, Neil and Veness, Joel and Desjardins, Guillaume and Rusu, Andrei A and Milan, Kieran and Quan, John and Ramalho, Tiago and Grabska-Barwinska, Agnieszka and others},
  journal={Proceedings of the national academy of sciences},
  volume={114},
  number={13},
  pages={3521--3526},
  year={2017},
  publisher={National Acad Sciences}
}

@article{wang2024survey,
  title={A survey on large language model based autonomous agents},
  author={Wang, Lei and Ma, Chen and Feng, Xueyang and Zhang, Zeyu and Yang, Hao and Zhang, Jingsen and Chen, Zhiyuan and Tang, Jiakai and Chen, Xu and Lin, Yankai and others},
  journal={Frontiers of Computer Science},
  volume={18},
  number={6},
  pages={186345},
  year={2024},
  publisher={Springer}
}

@article{shinn2024reflexion,
  title={Reflexion: Language agents with verbal reinforcement learning},
  author={Shinn, Noah and Cassano, Federico and Gopinath, Ashwin and Narasimhan, Karthik and Yao, Shunyu},
  journal={Advances in Neural Information Processing Systems},
  volume={36},
  year={2024}
}

@article{li2024evolving,
  title={Evolving agents: Interactive simulation of dynamic and diverse human personalities},
  author={Li, Jiale and Li, Jiayang and Chen, Jiahao and Li, Yifan and Wang, Shijie and Zhou, Hugo and Ye, Minjun and Su, Yunsheng},
  journal={arXiv preprint arXiv:2404.02718},
  year={2024}
}

@article{xu2022fast,
  title={Fast-lio2: Fast direct lidar-inertial odometry},
  author={Xu, Wei and Cai, Yixi and He, Dongjiao and Lin, Jiarong and Zhang, Fu},
  journal={IEEE Transactions on Robotics},
  volume={38},
  number={4},
  pages={2053--2073},
  year={2022},
  publisher={IEEE}
}

@misc{sun2025fastliohumanoid,
  author       = {Sun, Axiang and Zhu, Junfeng and Xu, Yihao and Li, Xin and Zhao, Zhongxia and Zhou, Gaohao and Li, Dongliang and An, Xiang and Tan, Huajie and Feng, Ziyong},
  title        = {{FAST\_LIO\_LOCALIZATION\_HUMANOID}},
  year         = {2025},
  howpublished = {Source code},
  url          = {https://github.com/deepglint/FAST_LIO_LOCALIZATION_HUMANOID},
  note         = {GitHub repository}
}

@article{yang2024efficient,
  title={Efficient global navigational planning in 3-d structures based on point cloud tomography},
  author={Yang, Bowen and Cheng, Jie and Xue, Bohuan and Jiao, Jianhao and Liu, Ming},
  journal={IEEE/ASME Transactions on Mechatronics},
  volume={30},
  number={1},
  pages={321--332},
  year={2024},
  publisher={IEEE}
}

@article{meng2025preserving,
  title={Preserving and combining knowledge in robotic lifelong reinforcement learning},
  author={Meng, Yuan and Bing, Zhenshan and Yao, Xiangtong and Chen, Kejia and Huang, Kai and Gao, Yang and Sun, Fuchun and Knoll, Alois},
  journal={Nature Machine Intelligence},
  pages={1--14},
  year={2025},
  publisher={Nature Publishing Group UK London}
}

@inproceedings{ouyang2025long,
  title={Long-horizon locomotion and manipulation on a quadrupedal robot with large language models},
  author={Ouyang, Yutao and Li, Jinhan and Li, Yunfei and Li, Zhongyu and Yu, Chao and Sreenath, Koushil and Wu, Yi},
  booktitle={2025 IEEE/RSJ International Conference on Intelligent Robots and Systems (IROS)},
  pages={11157--11164},
  year={2025},
  organization={IEEE}
}

@article{Bredche2017EmbodiedEI,
  title={Embodied Evolution in Collective Robotics: A Review},
  author={Nicolas Bred{\`e}che and Evert Haasdijk and Abraham Prieto},
  journal={Frontiers in Robotics and AI},
  year={2017},
  volume={5}
}

@article{adams2017formal,
  title={Formal definitions of unbounded evolution and innovation reveal universal mechanisms for open-ended evolution in dynamical systems},
  author={Adams, Alyssa and Zenil, Hector and Davies, Paul CW and Walker, Sara Imari},
  journal={Scientific reports},
  volume={7},
  number={1},
  pages={997},
  year={2017},
  publisher={Nature Publishing Group UK London}
}

@article{kunze2018longterm,
author = {Kunze, Lars and Hawes, Nick and Duckett, Tom and Hanheide, Marc and Krajník, Tomáš},
year = {2018},
month = {07},
pages = {1-1},
title = {Artificial Intelligence for Long-Term Robot Autonomy: A Survey},
volume = {PP},
journal = {IEEE Robotics and Automation Letters},
doi = {10.1109/LRA.2018.2860628}
}

@article{hawes2016strands,
author = {Hawes, N and Burbridge, C and Jovan, Ferdian and Kunze, Lars and Lacerda, Bruno and Mudrová, L and Young, J and Wyatt, Jeremy and Hebesberger, Denise and Koertner, Tobias and Ambrus, Rares and Bore, N and Folkesson, John and Jensfelt, Patric and Beyer, L and Hermans, Alexander and Leibe, B and Aldoma, A and Fäulhammer, Thomas and Hanheide, Marc},
year = {2016},
month = {10},
pages = {},
title = {The STRANDS Project: Long-Term Autonomy in Everyday Environments},
journal = {IEEE Robotics and Automation Magazine}
}

@article{Liu2021ALL,
  title={A Lifelong Learning Approach to Mobile Robot Navigation},
  author={Bo Liu and Xuesu Xiao and Peter Stone},
  journal={IEEE Robotics and Automation Letters},
  year={2021},
  volume={6},
  pages={1090-1096},
  url={https://api.semanticscholar.org/CorpusID:231704288}
}

@article{Hassanalian2018EvolutionOS,
  title={Evolution of space drones for planetary exploration: A review},
  author={Mostafa Hassanalian and Devyn Rice and Abdessattar Abdelkefi},
  journal={Progress in Aerospace Sciences},
  year={2018},
  volume={97},
  pages={61-105},
  url={https://api.semanticscholar.org/CorpusID:115274039}
}

@article{Chen2019Companionship,
author = {Chen, Na and Song, Jing and Li, Bin},
year = {2019},
month = {06},
pages = {1-7},
title = {Providing Aging Adults Social Robots’ Companionship in Home-Based Elder Care},
volume = {2019},
journal = {Journal of Healthcare Engineering},
doi = {10.1155/2019/2726837}
}

@article{Zeng2015autonomy,
author = {Zeng, Zheng and Lian, Lian and Sammut, Karl and He, Fangpo and Tang, Youhong and Lammas, Andrew},
year = {2015},
month = {12},
pages = {303-313},
title = {A survey on path planning for persistent autonomy of autonomous underwater vehicles},
volume = {110},
journal = {Ocean Engineering},
doi = {10.1016/j.oceaneng.2015.10.007}
}

\ifCLASSOPTIONcaptionsoff
  \newpage
\fi

\end{document}